\pdfoutput=1
\title{Thompson Sampling in Switching Environments with Bayesian Online Change Point Detection}
\author{
   Mellor, Joseph\\
   University of Manchester
   \and
   Shapiro, Jonathan\\
   University of Manchester
}
\documentclass[twoside]{article}

\usepackage{amsmath}
\usepackage{amsfonts}
\usepackage{algorithm}
\usepackage{algpseudocode}
\usepackage{graphicx}
\usepackage{url}
\usepackage[authoryear]{natbib}
\makeatletter
\renewcommand{\ALG@beginalgorithmic}{\small}
\makeatother

\begin{document}
 \maketitle



\begin{abstract}
Thompson Sampling has recently been shown to be optimal in the Bernoulli Multi-Armed Bandit setting\citep{Kaufmann2012}. This bandit problem assumes stationary distributions for the rewards. It is often unrealistic to model the real world as a stationary distribution. In this paper we derive and evaluate algorithms using Thompson Sampling for a Switching Multi-Armed Bandit Problem. 
\\
We propose a Thompson Sampling strategy equipped with a Bayesian change point mechanism to tackle this problem. We develop algorithms for a variety of cases with constant switching rate: when switching occurs all arms change (\emph{Global Switching}), switching occurs independently for each arm (\emph{Per-Arm Switching}), when the switching rate is known and when it must be inferred from data. This leads to a family of algorithms we collectively term \emph{Change-Point Thompson Sampling} (CTS).

 We show empirical results of the algorithm in 4 artificial environments, and 2 derived from real world data; news click-through\citep{yahoo} and foreign exchange data\citep{dukascopy}, comparing them to some other bandit algorithms. In real world data CTS is the most effective.
\end{abstract}
\section{Introduction} 

Thompson Sampling has recently been shown to be optimal in the Bernoulli Multi-Armed Bandit setting \citep{Kaufmann2012}. This bandit problem assumes stationary distributions for the rewards. It is often unrealistic to model the real world as a stationary distribution and algorithms such AdaptEvE have been proposed to solve bandit problems in this environment. In this paper we concern ourselves with a Switching Multi-Armed Bandit Problem. 

We first review Thompson Sampling, before describing the non-stationary environment we are concerned with. We then propose a method to solve this problem and review the techniques we employ. We then test our algorithms on a variety of environments.

\subsection{Thompson Sampling}

Thompson Sampling is effectively a probability matching algorithm. The desired strategy is to pull an arm with the probability that the arm is the best arm. This probability can be written as
\begin{equation}
P( a_i = a^* ) = \int_{\theta} I( a_i = a^* | \theta ) P( \theta | D ) d\theta ,
\label{eq:thompSamp}
\end{equation}
where \(\theta\) is a model of our arms, \(a^*\) is the optimal arm, \(a_i\) is the \(i^{\text{th}}\) arm and \(D\) is the history of past rewards. \(I(x)\) is the indicator function, which is \(1\) when \(x\) is true, and \(0\) otherwise.
The strategy can thus be reduced to sampling from the model distribution \(P(\theta | D )\) and then picking the arm that is maximal given this model. 

This paper considers multi-armed bandits with Bernoulli arms. Each arm, \(j\), delivers a reward of \(1\) with probability \(\theta_j\) and \(0\) otherwise. In the stationary case the parameter is \(\theta = ( \theta_1, \dots , \theta_k ) \) where \(k\) is the number of arms. The arms are assumed independent and so \(P(\theta | D )\) is a product of terms \(P(\theta_j | D_j )\), where \(D_j\) are past rewards for are \(j\). Further, since the arms are Bernoulli, for which the Beta distribution is a conjugate prior, we can write \(P(\theta)\) as a product of Beta distributions;
\begin{equation}
P(\theta | D ) = \prod_{j=1}^{k} P(\theta_j | \alpha_j,\beta_j, D_j)
\end{equation}
where \(\alpha_j = \text{\#\{reward = 1\}} + \alpha_0\) and \(\beta_j = \text{\#\{reward = 0\}} + \beta_0\).

We can sample from \(P(\theta | D )\) by sampling from all \(P(\theta_j| \alpha_j,\beta_j, D_j)\) and choosing the arm, \(j\), with largest \(\theta_j\).

\subsection{Model of Dynamic Environment}

In this paper we assume that the environment changes over time. We assume abrupt switching defined by a hazard function, \(h(t)\), such that,
\begin{equation}
\theta_i (t) = 
\left\{
\begin{array}{cc}
\theta_i (t-1) & \text{with probability}~h(t) \\
\theta_{\text{new}} \sim  U(0,1) & 1 - h(t).
\end{array}
\right.
\label{eq:switchProbs}
\end{equation}
The algorithms presented are designed with 2 such models in mind. The first model we will refer to as the \emph{Global Switching} model. This model switches at a constant rate, when a change point happens \emph{all} arms change their expected rewards. The second model will be referred to as \emph{Per-Arm Switching}. In this model change points occur independently for each arm, such that when the expected reward switches for one arm, it is uncorrelated to when all other arms switch. 

Examples of this form of changing environment might include stock market data where stock prices can change their statistical nature very quickly subject to external events. Switching behaviour has been studied in Financial Markets \citep{Preis2011}, and multi-armed bandits have been applied to this field before \citep{Sorensen2007}.
\subsection{Regret for Switching Environments}
In a stationary multi-armed bandit the regret measure we use is taken to be the expected difference in reward between our strategy and that of a policy which always chose the arm with highest expected payoff, \(\mu^*\).
In a switching system the arm which is considered optimal changes. In this case a different form of regret is often used. \citet{Garivier2008} use the following definition of regret.
\begin{equation}
R(T) = < \sum_{t=0}^T (\mu_t^* - \mu_{i_t}) >,
\label{eq:nonStatRegret}
\end{equation}
where \(\mu_t^*\) is the highest expected payoff of an arm at time \(t\) and  where \(<.>\) denotes expectation over possible sequences of \(\{\mu_{i_t}\}\).
\\
Unless otherwise stated our experiments will report a related quantity,
\begin{equation}
R_n(T) = < \sum_{t=0}^T I( \mu_t^* \neq \mu_{i_t} ) > ,
\label{eq:experRegret}
\end{equation}
which is the expected number of times a suboptimal arm is pulled. This corresponds to the results \citet{Hartland2006} report.
\section{Switching Thompson Sampling}

In order to perform Thompson Sampling we wish to sample from \(P(\theta | D_{t-1})\), which is the probability of the arm model given the data so far. In a switching system the arms model \(\theta\) is only dependent on the data since the last switching occurred, but we do not know when this happened. If we did we could just do the same Bayesian update as with the standard Bernoulli case to arrive at the distribution of our model. Since we do not know the runlength \(r_t\) we can introduce it as a latent variable and marginalise it out. Taking \(D_{t-1}\) as the history of rewards and arm pulls seen so far, we can write this as
\begin{equation}
P( \theta | D_{t-1}) = \sum_{r_t} \underbrace{P(\theta | D_{t-1}, r_t)}_{\text{\parbox{2.5cm}{\begin{center}posterior of \\[-4pt] model given data\end{center}}}} \overbrace{P( r_t | D_{t-1})}^{ \text{\parbox{2cm}{\begin{center}probability \\[-4pt] of runlength\end{center}}} }.
\label{eq:switchPoint}
\end{equation}
\\
Now to sample from \(P(\theta | D_{t-1})\) we just need to sample from the \(P(r_t | D_{t-1})\) (the runlength distribution) and then given that runlength, sample from \(P(\theta | D_{t-1}, r_t)\) to arrive at our arm model \(\theta\). We can select the arm to pull that in expectation maximises the reward given this model. 

\section{Bayesian Online Change Detection}
\citet{Fearnhead2007} as well as \citet{Adams2007} have independently done work on calculating the online posterior of the runlength. They show exact inference on the runlength can be achieved by a simple message passing algorithm. Letting \(x_t\) be the reward at time \(t\) so that \(D_t = x_t \cup D_{t-1}\).
The inference procedure can be easily derived as follows.
\begin{equation}
P(r_t | x_{t-1}, D_{t-2}) = \frac{P(r_t,x_{t-1},D_{t-2})}{P(x_{t-1},D_{t-2})}
\label{eq:runlen1}
\end{equation}
The numerator can then be expressed as
\begin{align}
P(r_t,x_{t-1},D_{t-2}) \nonumber\\ &\hspace{-2.5cm}=\sum_{r_{t-1}} P(r_t,r_{t-1},x_{t-1},D_{t-2}) \\
&\hspace{-2.5cm}= \sum_{r_{t-1}}  P(r_t,x_{t-1} | r_{t-1}, D_{t-2}) P( r_{t-1}, D_{t-2}) \label{eq:messPass1}\\
&\hspace{-2.5cm}= \sum_{r_{t-1}} \overbrace{P(r_t | r_{t-1})}^{\text{switching rate}} \overbrace{P( x_{t-1} | r_{t-1}, D_{t-2} )}^{\text{reward likelihood}} P( r_{t-1},D_{t-2}).\label{eq:messPass2}
\end{align}
\\
The derivation just applies the rules of probability up to and including equation \ref{eq:messPass1}. One assumption is made in equation \ref{eq:messPass2}, that the runlength is only dependent on the previous time steps runlength. This forms a simple message passing algorithm because \(r_t\) can only take values depending on \(r_{t-1}\). In fact \(r_t = r_{t-1}+1\) when switching does not occur and \(r_t=0\) when it does. 
\(P(r_t | r_{t-1})\) is defined by a hazard function \(h(t)\). For simplicity in our case this is a constant switching rate \(\gamma\).


Unfortunately the exact inference has space and time requirements that grow linearly in time. The space requirements are linear because at each time step the support set of the posterior runlength distribution increases by one, which means we have to store information for an extra value of the runlength at every step. The update is also linear in time, as the message passing algorithm requires an update to each runlength in the support. Adams and MacKay suggest a simple thresholding technique to eliminate runlengths with small probability mass associated with them. As we can only know in expectation how much memory this algorithm will require, an alternative with hard guarantees on memory requirements is desirable. Fearnhead and Liu suggest a much more sophisticated particle filter resampling step to maintain a finite sample of the runlength distribution, which has the benefit that we can be certain on the upper limit of space the algorithm requires, this approach is the one taken in this paper. 

\subsection{Particle Filters}

A particle filter is a Monte-Carlo method for approximately estimating a sequential Bayesian model. Particles are used to represent points in the distribution to be estimated and are assigned weights that correspond to their approximate probabilities. The number of particles can grow at each time step and so occasionally some particles need to be thrown away. This leaves us to assign new weights to the remaining particles. This procedure is called resampling.

\subsubsection{Stratified Optimal Resampling}

\citet{Fearnhead2003} originally proposed optimal resampling. We wish to reduce a discrete probability distribution with a support of \(N\) discrete points down to a stochastic distribution of \(M\) discrete points, where the set of \(M\) points is a subset of the original \(N\). The original \(N\) points each have probability mass \(p_i\) associated with them, and the procedure finds a reweighting of these probabilities, \(q_i\) such that \(N-M\) of the probabilities are \(0\). The idea is that we wish there to be no bias in the sampling procedure, which means that the expected value of \(q_i\) should be the original probability mass \(p_i\). The algorithm is optimal in the sense that the expected squared difference between the original probabilities \(p_i\) and the new weights \(q_i\) are minimised. 

This can be done by the following procedure;

\begin{enumerate}
\item Find \(\kappa\) such that  \(M = \sum_{i=1}^N \min ( 1, p_i/\kappa)\)
\item Sample \(u\) from uniform distribution, \(U(0,\kappa)\)
\item Iterate through all \(p_i\)
\begin{enumerate}
\item If \(p_i > \kappa\) Then \(q_i = p_i\)
\item Otherwise 
\begin{enumerate}
\item \(u = u - p_i\)
\item If \(u < 0\) Then \(q_i = \kappa\) and \( u = u + \kappa\) 
\item Otherwise \(q_i = 0\) 
\end{enumerate}
\end{enumerate}
\end{enumerate}

The particles where \(p_i = q_i\) are kept with probability \(1\). The remaining particles are such that \(q_i=\kappa\) with probability \(p_i/\kappa\) and \(q_i=0\) otherwise. Thus their expectation remains the same.

The worstcase time complexity of this algorithm is \(O(N \log N )\), but it has an amortised cost of \(O(N)\) \citep{Fearnhead2003}. 

\section{Proposed Inference Models}
We have shown we can perform Thompson Sampling in a switching system by splitting the procedure into a stage that samples the runlength since a switch occurred and a stage that samples from the arm model given this runlength. 

The Global Switching and Per-Arm Switching models are appealing due to their simplicity. Only one runlength distribution needs to be inferred for Global Switching, which does not depend on the number of arms the bandit has. The Per-Arm model can store the runlengths for each arm independent, and the space requirements grow linearly with respect to the number of arms. Other models with more complicated dependencies between arms can quickly become intractable.
\begin{algorithm}[tb!]
\caption{Global Change-Point Thompson Sampling}\label{alg:OCTSAlg}
\begin{algorithmic}
\scriptsize
\Procedure{Global-CTS}{$N,\gamma,\alpha_0=1,\beta_0=1$}
\State $t\gets 0$ \Comment{Initialise time}
\State $w_0^t \gets 1$, and add to \(\{w\}^t\) \Comment{Initialise runlength distribution}
\State For all arms $j$, $\alpha_{0,j}^t \gets \alpha_0$ \Comment{Initialise hyperparameters}
\State For all arms $j$, $\beta_{0,j}^t \gets \beta_0$
\While{Interacting}
\State $a\gets \Call{SelectAction}{\{w\}^t,\{\alpha\}^t,\{\beta\}^t}$
\State $r\gets \Call{PullArm}{a}$
\State $\{w\}^{t+1}\gets\Call{UpdateChangeModel}{\{w\}^t,\{\alpha\}_a^t,\{\beta\}_a^t,a,r},\gamma)$
\State $\{\alpha\}^t,\{\beta\}^t\gets\Call{UpdateArmModels}{\{\alpha\}^t,\{\beta\}^t,a,r}$
\If {$\left|\{w\}^{t+1}\right| = N$ }
\State $\Call{ParticleResample}{\{w\}^{t+1},\{\alpha\}^{t+1},\{\beta\}^{t+1}}$ 
\EndIf
\State $t\gets t + 1$
\EndWhile
\EndProcedure
\\\hrulefill
\Procedure{UpdateChangeModel}{$\{w\}^t,\{\alpha\}_a^t,\{\beta\}_a^t,a,r,\gamma$}
\If{$r=1$}
\State $likelihood_i\gets \frac{\alpha_{i,a}^t}{\alpha_{i,a}^t + \beta_{i,a}^t}$,    For all \(i\) s.t. \(w_i^t \in \{w\}^t\)
\Else
\State $likelihood_i\gets \frac{\beta_{i,a}^t}{\alpha_{i,a}^t + \beta_{i,a}^t}$,    For all \(i\) s.t. \(w_i^t \in \{w\}^t\)
\EndIf
\State $w_{i+1}^{t+1}\gets (1-\gamma)*likelihood_i*w_i^t $,    For all \(i\) s.t. \(w_i^t \in \{w\}^t\)
\State $w_{0}^{t+1}\gets \sum_{i} \gamma*likelihood_i*w_i^t $
\State Normalise $\{w\}^{t+1}$
\State  $\textbf{return} \{w\}^{t+1}$
\EndProcedure
\\\hrulefill
\Procedure{UpdateArmModels}{$\{\alpha\}^t,\{\beta\}^t,a,r$}
\If{r=1}
\State $\alpha_{i+1,a}^{t+1}\gets \alpha_{i,a}^{t} + 1$,  For all \(i\) s.t. \(\alpha_{i,a}^t \in \{\alpha\}_a^t\)
\Else
\State $\beta_{i+1,a}^{t+1}\gets \beta_{i,a}^{t} + 1$,  For all \(i\) s.t. \(\beta_{i,a}^t \in \{\beta\}_a^t\)
\EndIf
\State $\alpha_{0,j}^{t+1}\gets \alpha_0$ ,  For all arms \(j\) \Comment{Set Prior for runlength 0}
\State $\beta_{0,j}^{t+1}\gets \beta_0$ ,  For all arms \(j\)
\State  $\textbf{return} \{\alpha\}^{t+1},\{\beta\}^{t+1}$
\EndProcedure
\Procedure{ParticleResample}{$\{w\}^{t+1},\{\alpha\}^{t+1},\{\beta\}^{t+1}$}
\State Find set to discard \(d\in D\) using Stratified Optimal Resampling on \(\{w\}^{t+1}\)
\State Discard all \(w_d^{t+1}\), \(\alpha_d^{t+1}\), \(\beta_d^{t+1}\) 
\EndProcedure
\\\hrulefill
\Procedure{SelectAction}{$\{w\}^t,\{\alpha\}^t,\{\beta\}^t$}
\State Pick \(i\) with probability \(w_i^t\)
\For { each arm \(j\) }
\State $sample_j \gets Beta(\alpha_{i,j}^{t},\beta_{i,j}^{t})$
\EndFor
\State 	$\textbf{return} \max_j sample_j $
\EndProcedure
    \end{algorithmic}
  \end{algorithm}

\subsection{Global switching}

In global switching there is a single change point process across all of the arms since when one arm switches distribution so do all other arms. This means that the data from every arm pull contributes to the posterior of the single runlength distribution. Effectively to sample from the posterior of the full bandit model, we first need to sample from the runlength distribution, this gives us an estimate of the runlength, which tells us how much data from the past our arms can use. Once the global runlength is sampled, we then proceed by sampling individually from the posterior distributions of the arms, given only the data since the last changepoint (determined by the runlength). The arm with the corresponding maximum sample is then pulled. We only need to store the posterior probabilities of the given runlengths and the hyperparameters for the arm posteriors associated with those runlengths. We will call the runlength distribution the Change Point model, and the set of hyperparameters associated with each runlength for a given action the Arm model. 
\\
The Change Point model is an approximation of the runlength distribution storing a probability for at most \(N\) runlengths. Let \(w_i^t\) be the probability of having a runlength of \(i\) at time \(t\). In this paper the arm rewards are assumed to come from a Bernoulli distribution so the hyperparameters stored are the 2 parameters for the Beta distribution. Let \(\alpha_{i,j}^t\) and \(\beta_{i,j}^t\) be the hyperparameters for a runlength of \(i\) at time \(t\) for arm \(j\). At any point in time \(t\) there is a set of runlengths \(R_t \subset \mathbb{N}\), \(\left|R_t\right| \leq N\), where for every \(r \in R_t\) there exists quantities \(w_r^t\), \(\alpha_{r,j}^t\) and \(\beta_{r,j}^t\). When \(\left|R_t\right|=N\) then a resampling step is performed in order to reduce the number of runlengths stored. For ease of notation let \(\{w\}^t\) be the set of runlength probabilities at time \(t\) and let \(\{\alpha\}_j^t\) and \(\{\beta\}_j^t\) be the sets of hyperparameters for arm \(j\) at time \(t\). Similarly let \(\{\alpha\}^t\) and \(\{\beta\}^t\) be the set of all hyperparameters at time \(t\). The algorithm is presented in pseudocode in figure \ref{alg:OCTSAlg}.

We will refer to this algorithm as \emph{Global Change-Point Thompson Sampling } (Global-CTS).


\subsection{Per-arm switching}

The difference in implementation with respect to global switching is that now there is a runlength distribution for each arm. That is, for each arm \(j\) we have a different set of runlength probabilities \(w_{i,j}^t \in \{w\}_j^t\). In the per-arm switching model at a timestep \(t\) we update the Change Point model associated with the arm that was pulled at \(t\) much like via the update equations sketched in \ref{eq:messPass2}.
\\
The Change Point models associated with arms not pulled at \(t\) are updated differently since the runlength for these arms is independent of the reward we received for the arm we actually pulled.
The reward likelihood term disappears in the update equations for the runlength distribution of unpulled arms. This is shown in equation \ref{eq:indmessPass}. Since we normalise the distribution at each step we can ignore the factor \(P(x_{t-1})\). This is shown as follows,
\begin{align}
P(r_t,x_{t-1},D_{t-2}) &= P(x_{t-1}) P(r_t,x_{t-1},D_{t-2}) \\
&\propto \sum_{r_{t-1}} P(r_t,r_{t-1},D_{t-2}) \\
&\propto \sum_{r_{t-1}} P(r_t | r_{t-1}, D_{t-2}) P( r_{t-1}, D_{t-2}).
\label{eq:indmessPass}
\end{align}

We will refer to this algorithm as \emph{Per-Arm Change-Point Thompson Sampling} (PA-CTS).

\section{Learning the Switching Rate}
Both \citet{Wilson2010} and \citet{Turner2009} have proposed methods for learning the hazard function from the data. Wilson et al. method can learn a hazard function that is piecewise constant via a hierarchical generative model. Turner et al. can learn any parametric hazard rate via gradient descent, but from initial investigations appeared to not perform particularly well if the hazard rate is adapted at every time step. For the purposes of this paper, a constant switching rate was assumed which was learned using the approach of Wilson et al.
\\
For the simplest case where we consider a single constant switch rate, Wilson et al. model whether a change point occurred as a Bernoulli variable. The hyperparameters of this switching rate are those of a Beta distribution and can be thought of as the number of times the system has switched, \(a_t\) and the number of times it not switched \(b_t\). We now compute the joint distribution \(P(r_t,a_t | x_{t-1}, D_{t-2})\) as oppose to the original distribution \(P(r_t| x_{t-1}, D_{t-2})\). The message passing proceeds in a very similar fashion as before, except now the number of particles also grows quadratically rather than linearly.
\\
In the global switching model the algorithm now keeps track of sets of particles \( w_{r,a}^t \), \( \alpha_{r,i,a}^t \) and \( \beta_{r,i,a}^t \) associated with a runlenght \(r\), learning rate hyperparameter \(a\), arm \(i\) and time \(t\). The updates are as follows.
\begin{align*}
w_{0,0}^0 &\gets 1 \\
w_{r+1,a}^{t+1} &\gets \frac{t-a+1}{t+2} \frac{\alpha_{r,i,a}^t}{\alpha_{r,i,a}^t + \beta_{r,i,a}^t} w_{r,a}^{t} \text{if reward}= 1 \\
w_{r+1,a}^{t+1} &\gets \frac{t-a+1}{t+2} \frac{\beta_{r,i,a}^t}{\alpha_{r,i,a}^t + \beta_{r,i,a}^t} w_{r,a}^{t} \text{if reward}= 0 \\
w_{r,a+1}^{t+1} &\gets \frac{a+1}{t+2} \frac{\alpha_{r,i,a}^t}{\alpha_{r,i,a}^t + \beta_{r,i,a}^t} w_{r,a}^{t} \text{if reward}= 1 \\
w_{r,a+1}^{t+1} &\gets \frac{a+1}{t+2} \frac{\beta_{r,i,a}^t}{\alpha_{r,i,a}^t + \beta_{r,i,a}^t} w_{r,a}^{t} \text{if reward}= 0 \\
\end{align*}
We again use the resampling algorithm of Fearnhead to manage the space requirements of the algorithm. 


In the Global Switching model there is only 1 runlength distribution, and so only 1 switching rate to learn, this leads naturally to an algorithm \emph{Non-Parametric Global Change-Point Thompson Sampling} (NP Global-CTS). With Per-Arms there are many possibilities, there could be a single switching rate for each of the independent arms, or each arm could have a separate switching rate. In this paper we assume each arm has a separate switching rate and call this algorithm \emph{Non-Parametric Per-Arm Change-Point Thompson Sampling} (NP PA-CTS).


\section{Tracking Changes In The Best Arm}

The algorithms presented so far attempt to track changes in all arms, irrespective of whether they are pulled. The distribution of \(\theta_i\) for an arm not pulled will slowly become flatter and thus have higher variance. One of the stated assumptions of Adapt-EvE, was that it was only important to track whether the perceived best arm has changed distribution. We can modify the algorithms to better replicate this assumption. 
\\
The model for the Per-Arm Switching is adapted most simply. Since both the Change Point models and the Arm models are independent for each arm, we can track the change of the best arm by only updating the Change Point and Arm models for the arm that was pulled. For the Global Switching model the change is not so clear since all arms share a Change Point model. In this paper we updated the shared Change Point model and only updated the Arm model for the arm pulled. The question then arises what should the hyperparameters for the unpulled arms be that are associated with the new runlength of zero. The approach taken here was to set them to the hyperparameters associated with a random other runlength in the distribution. 
\\
We can apply the sample method from Wilson to infer the switching rate for these architectures as well. The algorithms described in this section with be denoted by having a ``2'' appended to the algorithm name.

\section{Experiments}

6 different non-stationary environments were used to evaluate our bandit model. 4 are based on purely synthetic data, and 2 use data collected from the real world. The parameters for all experiments shown were tuned based on the PASCAL challenge. Bold denotes best results in all tables. 

\subsection{Global Switching Environment}
We first compare the algorithms in an environment with a constant global switching rate. Global-CTS and NP Global-CTS were designed for this environment and so a-priori we would expect them to perform the best. 

\begin{figure}[h!]
\centering
\includegraphics[width=0.8\textwidth]{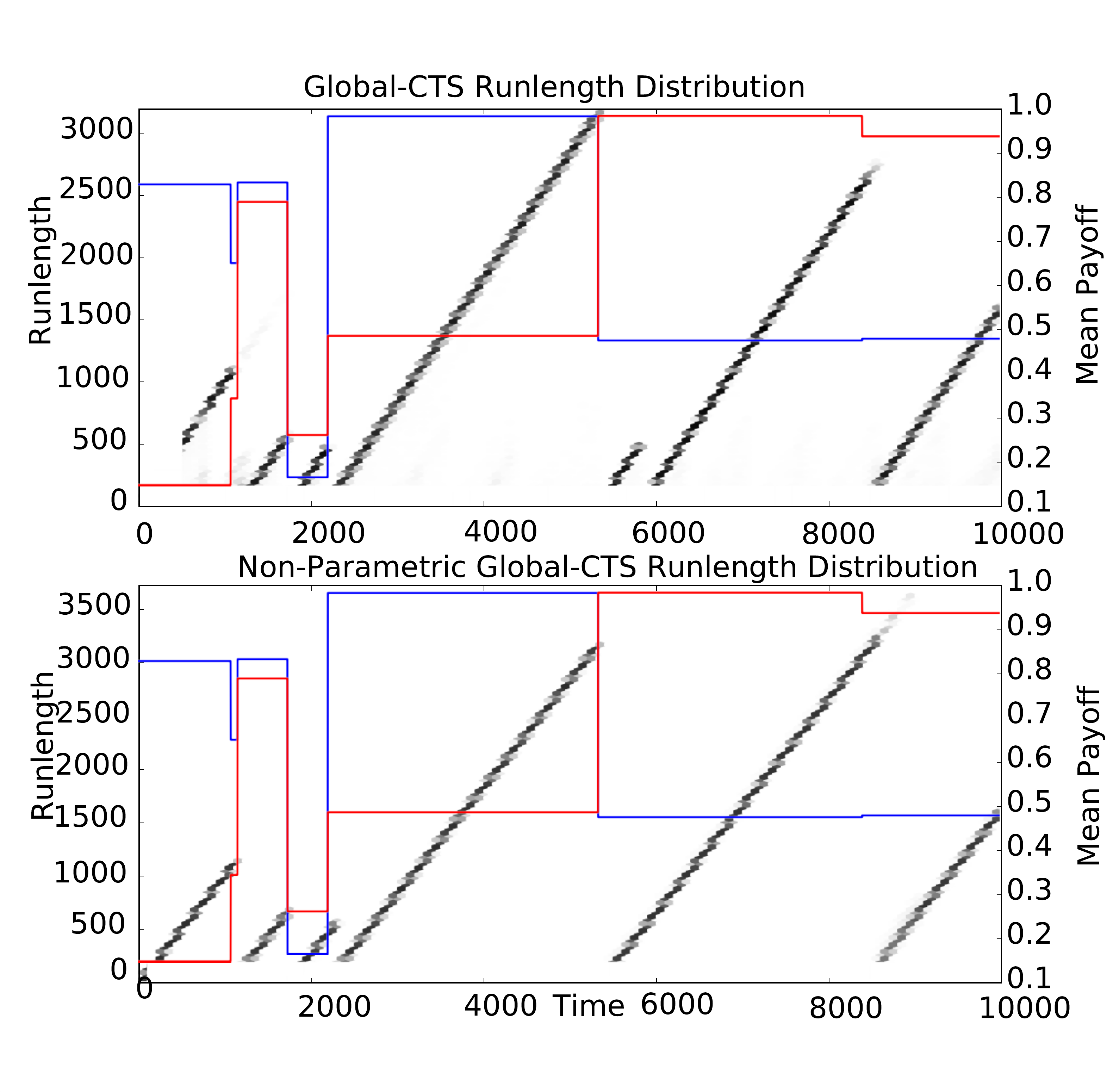}
\caption{Runlength Distribution for Global-CTS and NP Global-CTS in Global Switching Environment. The mean payoffs of the arms are super-imposed over the distribution.}
\label{fig:globalRunPost1}
\end{figure}

\begin{figure}[h!]
\centering
\includegraphics[width=0.8\textwidth]{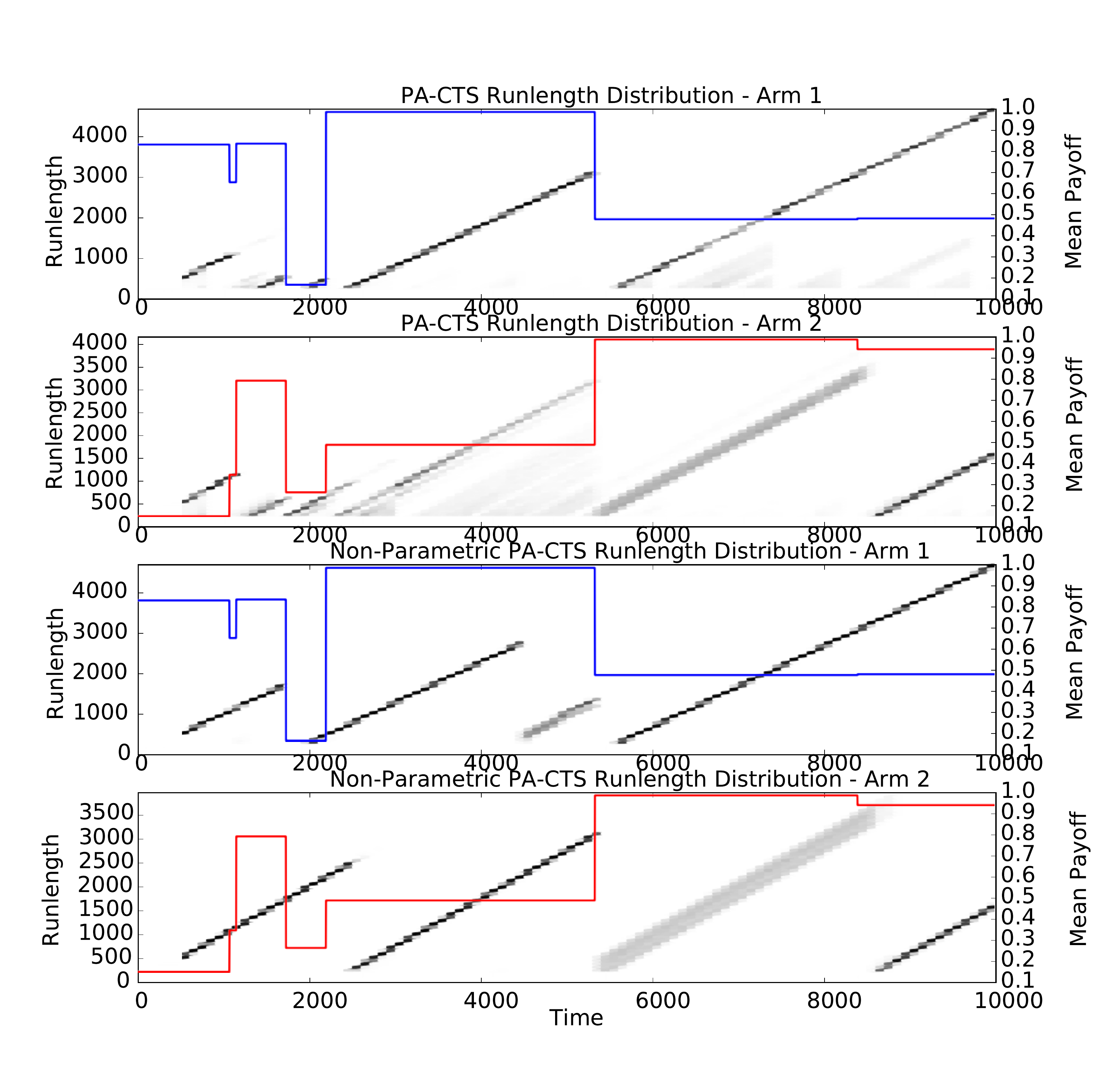}
\caption{Runlength Distribution for PA-CTS and NP PA-CTS in Global Switching Environment. The mean payoffs of the arms are super-imposed over the distribution.}
\label{fig:globalRunPost3}
\end{figure}

The first set of experiments were a single run of the algorithms working in an instance of this environment type with 2 arms.
 Figures \ref{fig:globalRunPost1} and \ref{fig:globalRunPost3} plot example heatmaps of the runlength distributions of some of the algorithms. At a particular time, the graphs show the runlength distribution. In the case of the PA-CTS and NP PA-CTS there are 2 plots for each algorithm, corresponding to the runlength distribution for each arm. The pay off of the 2 arms has been superimposed over the top of the plots so that it can be seen how the runlength distribution matches up with the changes in the environment
From the heatmap figures we can see the change point prediction works when applied to a bandit problem. As expected the change point distribution looks to be more accurate for the Global-CTS and NP Global-CTS algorithms which use the Global Switching model, this is because each data point can contribute to the posterior runlength distribution. The PA-CTS also performs reasonably well even though the amount of data that has influence on each posterior is reduced. For the NP PA-CTS algorithm, learning the separate switching rates appears to significantly decrease the certainty for a particular runlength. 
%

~


An experiment comparing the algorithms in this setting was performed. Each run was over a period of \(10^6\) time steps and the experiment was repeated 100 times. The results are displayed in table \ref{tbl:Global}. All parameters were set as for the PASCAL challenge test run. The environment constant switching rate was \(10^{-4}\).
\\
Global-CTS performs the best in the environment, which is not surprising since the environment fits the algorithms model. NP Global-CTS performs well in this too, which suggests that learning the hazard rate for this model may be feasible. 

\begin{table}[ht]
\centering
\scriptsize
\setlength{\tabcolsep}{3pt}
\caption{Results against Global Switching Environment (given as number of mistakes\(\times 10^{-3} \pm\) Std. Error)}
\begin{tabular}{l r | l r }
 Name & Regret & Name & Regret \\
\hline\hline\hline
 \multicolumn{1}{|| l}{ Global-CTS  }      & \(\mathbf{5.9\pm0.07}\)   & Global-CTS2   &  \multicolumn{1}{r ||}{ \(30.5\pm1.07\) }   \\
 \multicolumn{1}{|| l}{ PA-CTS }           & \(12.1\pm0.10\)           & PA-CTS2       &  \multicolumn{1}{r ||}{ \(49.6\pm1.70\)  } \\
 \multicolumn{1}{|| l}{ NP Global-CTS  }   & \(6.7\pm0.08\)            & NP PA-CTS     & \multicolumn{1}{r ||}{  \(29.4\pm0.95\)  }  \\
 \multicolumn{1}{|| l}{ NP Global-CTS2  }  & \(10.3\pm0.20\)           & NP PA-CTS2    &  \multicolumn{1}{r ||}{ \(25.6\pm0.86\)  }  \\
\hline
 UCB               & \(178.3\pm8.20\)         & DiscountedUCB & \(15.5\pm0.27\)    \\
 Random            & \(333.1\pm2.09\)         &               &                   \\
\end{tabular}
\label{tbl:Global}
\end{table}

\subsection{Per-Arm Switching Environment}
The next environment was a switching system where the switching for each arm was independent of every other arm. PA-CTS and NP PA-CTS were designed with this situation in mind and again a-priori may be expected to perform better.
\\




%
%
An experiment comparing the algorithms was performed with \(10^6\) iterations and then repeated \(100\) times. The results are shown in table \ref{tbl:PerArm}.
 As expected the PA-CTS algorithm performs best in this environment. NP PA-CTS, the algorithm corresponding to PA-CTS that learns the hazard rate suffers much more regret, which would appear to indicate for the particular model the parameters are not being learned quickly enough. The algorithms designed for a Global Switching environment also perform reasonably in this sort of environment.
\\

\begin{table}[ht]
\centering
\scriptsize
\setlength{\tabcolsep}{2pt}
\caption{Results against Per-Arm Switching Environment (given as number of mistakes\(\times 10^{-3} \pm\) Std. Error)}
\begin{tabular}{l r | l r }
Name & Regret & Name & Regret \\
\hline\hline
 \multicolumn{1}{|| l}{ Global-CTS     }  & \(13.8\pm0.20\)          & Global-CTS2   &  \multicolumn{1}{r ||}{ \(37.9\pm1.02\) } \\
 \multicolumn{1}{|| l}{ PA-CTS         }  & \(\mathbf{13.0\pm0.11}\) & PA-CTS2       &  \multicolumn{1}{r ||}{ \(67.1\pm1.23\) } \\
 \multicolumn{1}{|| l}{ NP Global-CTS  }  & \(13.8\pm0.17\)          & NP PA-CTS     &  \multicolumn{1}{r ||}{ \(30.8\pm0.79\) } \\
 \multicolumn{1}{|| l}{ NP Global-CTS2 }  & \(15.8\pm0.28\)          & NP PA-CTS2    &  \multicolumn{1}{r ||}{ \(38.1\pm0.83\) } \\
\hline
UCB              & \(175.1\pm7.47\)        & DiscountedUCB & \(16.8\pm0.28\)  \\
Random           & \(336.4\pm1.85\)        &               &                 \\
\end{tabular}
\label{tbl:PerArm}
\end{table}

\subsection{Bernoulli Armed Bandit with Random Normal Walk}
The PASCAL challenge environments were found to be periodic, which is not the sort of environment our algorithms were intended for. Another simulated environment was investigated. In this environment at time,\(t\), each arm, \(i\), was Bernoulli with probability of success \(\theta_i(t)\). At each time step the success rate of the arm was allowed to drift as a truncated normal walk. 
That is the probability of success for an arm \(\theta_i(t) \in [0,1]\) conditional on \(\theta_i(t-1) \in [0,1] \) is,
\begin{equation}
P(\theta_i(t) | \theta_i(t-1)) = \frac{e^{\frac{-(\theta_i(t-1)-\theta_i(t))^2}{\sigma^2}}}{\sqrt{2\pi}\sigma \int_0^1 \frac{e^{\frac{-(\theta_i(t-1)-x)^2}{\sigma^2}}}{\sqrt{2\pi}\sigma}dx }.
\label{eq:truncNormal}
\end{equation}

Table \ref{tbl:Trunc} shows an comparison of the algorithms. The experiment was run \(100\) times, where each run had a period of \(10^6\). The variance of the random walk was set to \(\sigma^2=0.03\). 

\begin{table}[ht]
\centering
\scriptsize
\setlength{\tabcolsep}{3pt}
\caption{Results against Bernoulli Bandit with Truncated Normal Walk (given as number of mistakes\(\times 10^{-3} \pm\) Std. Error)}
\begin{tabular}{l r | l r }
Name & Regret & Name & Regret \\
\hline\hline
 \multicolumn{1}{|| l}{ Global-CTS     }   & \(97.9\pm0.10\)   & Global-CTS2     &  \multicolumn{1}{r ||}{ \(134.1\pm0.23\)            }   \\
 \multicolumn{1}{|| l}{ PA-CTS         }   & \(107.1\pm0.16\)  & PA-CTS2         &  \multicolumn{1}{r ||}{ \(148.9\pm0.35\)            }   \\
 \multicolumn{1}{|| l}{ NP Global-CTS  }   & \(116.6\pm0.13\)  & NP PA-CTS       &  \multicolumn{1}{r ||}{ \(117.0\pm0.11\)            }   \\
 \multicolumn{1}{|| l}{ NP Global-CTS2 }   & \(100.9\pm0.10\)  & NP PA-CTS2      &  \multicolumn{1}{r ||}{ \(\mathbf{94.8\pm0.13}\)    }   \\
\hline
UCB               & \(194.5\pm3.78\)  & DiscountedUCB   & \(162.4\pm0.47\)               \\
Random            & \(325.9\pm0.24\)  &                 &                               \\
\end{tabular}
\label{tbl:Trunc}
\end{table}

In this sort of environment it appears that our algorithms perform better than the benchmark algorithms with NP PA-CTS2 achieving the smallest regret.
\subsection{PASCAL Challenge 2006}
The PASCAL Exploration vs. Exploitation Challenge 2006 was a competition in a multi-armed bandit problem\citep{PASCAL}. The challenge revolved around website content optimisation, whereby the options available corresponded to different content to present to a user on a website. The challenge is a good general test for the algorithms presented in this paper as to perform well it was required for the bandit algorithms to be able to work in non-stationary environments. 
The challenge had 6 separate environments in which the algorithms needed to perform; Frequent Swap (FS), Long Gaussians (LG), Weekly Variation (WV), Daily Variation (DV), Weekly Close Variation (WCV) and Constant (C). These environments are artificially generated, where the dynamics of the expected payoffs resemble either periodic Gaussian, Sinusoidal or constant signals. 
\\
\citet{Hartland2006} won this competition with the Adapt-EvE algorithm. The Adapt-EvE algorithms most prominent feature is its use of a change-point detection mechanism. Since the algorithms presented in this paper also use a change-point mechanism it is interesting to compare their performance. The challenge also provides an environment for which the algorithm was not directly designed for and so will hopefully indicate some robustness in their strategy. We were unable to implement a version of Adapt-EvE that replicated the performance reported, so here we are simply replicating the results published. To avoid an unfair comparison in other environments we did not run our own implementation of Adapt-EvE in those environments. 

Table \ref{tbl:Pascal2006} shows a comparison of the Change-Point Thompson Sampling algorithms (Global-CTS, PA-CTS, NP Global-CTS NP PA-CTS) against Adapt-EvE Meta-Bandit and Meta-p-Bandit \citep{Hartland2006}. The comparison also features the algorithm ``DiscountedUCB'',  which was submitted by Thomas Jaksch to the same competition and performed comparably to Adapt-EvE. The code for this algorithm was available and so has been included for comparison in all other environments.

\begin{table}[ht]
\centering
\footnotesize
\scriptsize
\setlength{\tabcolsep}{3pt}
\caption{Results against PASCAL EvE Challenge 2006 (given as number of mistakes\(\times 10^{-3}\))}
\begin{tabular}{l ||r r|| r }
\hline\hline
      &   Global-CTS                     &     Global-CTS2            &      Adapt-EvE Meta $\rho$         \\
\hline                                                                                            
WCV   &   \(8.9\pm0.4\)                  &   \(6.9\pm0.4\)            &      \(5.5\pm0.9\)               \\
FS    &   \(27.9\pm2.4\)                 &   \(12.5\pm1.3\)           &      \(10.6\pm1.3\)              \\
C     &   \(\mathbf{0.6\pm0.1}\)         &   \(1.0\pm0.2\)            &      \(3.2\pm0.3\)               \\
DV    &   \(17.1\pm0.3\)                 &   \(6.6\pm0.3\)            &      \(6.1\pm0.7\)               \\
LG    &   \(4.4\pm0.4\)                  &   \(3.4\pm0.5\)            &      \(4.3\pm1.4\)               \\
WV    &   \(8.2\pm0.3\)                  &   \(5.3\pm0.5\)            &      \(5.1\pm0.9\)               \\
\hline                                                                                            
Total &   67.2                           &   35.8                     &      \textbf{34.7}                \\
\hline
      & PA-CTS                    & PA-CTS2                     &    Adapt-EvE Meta     \\
\hline                                                                              
WCV   & \(\mathbf{4.2\pm0.8}\)    & \(6.2\pm0.4\)               &    \(5.4\pm0.8\)     \\
FS    & \(13.7\pm1.6\)            & \(15.1\pm1.7\)              &    \(14.0\pm1.9\)        \\
C     & \(3.2\pm0.4\)             & \(2.0\pm0.3\)               &    \(2.5\pm0.5\)          \\
DV    & \(\mathbf{4.5\pm1.5}\)    & \(4.9\pm0.5\)               &    \(6.2\pm0.7\)          \\
LG    & \(9.4\pm2.9\)             & \(3.7\pm0.7\)               &    \(4.8\pm1.6\)          \\
WV    & \(4.7\pm1.7\)             & \(5.4\pm0.5\)               &    \(4.8\pm0.8\)          \\
\hline                                                                              
Total & 39.6                      & 37.4                        &    37.7                   \\
\hline
      &  NP Global-CTS             &   NP Global-CTS2     &  DiscountedUCB              \\
\hline                                                                               
WCV   &  \(8.9\pm0.4\)             &    \(9.0\pm0.3\)     &  \(5.3\pm0.5\)            \\
FS    &  \(28.2\pm2.7\)            &    \(14.8\pm1.2\)    &  \(\mathbf{10.1\pm1.1}\)    \\
C     &  \(0.3\pm0.2\)             &    \(0.8\pm0.2\)     &  \(5.5\pm0.5\)              \\
DV    &  \(17.6\pm0.3\)            &    \(16.0\pm0.3\)    &  \(7.9\pm0.9\)             \\
LG    &  \(4.4\pm0.4\)             &    \(4.0\pm0.3\)     &  \(\mathbf{2.9\pm0.4}\)     \\
WV    &  \(8.5\pm0.3\)             &    \(8.4\pm0.3\)     &  \(\mathbf{4.0\pm0.4}\)     \\
\hline                                                                               
Total &  67.9                      &  53.1                &  35.7                        \\
\hline
      & NP PA-CTS                 & NP PA-CTS2                & Random          \\
\hline                                                                          
WCV   & \(12.8\pm0.7\)            & \(10.4\pm0.4\)            & \(25.7\pm0.3\)   \\
FS    & \(23.1\pm1.2\)            & \(23.0\pm1.9\)            & \(49.1\pm0.5\)   \\
C     & \(15.8\pm0.4\)            & \(1.9\pm0.2\)             & \(20.0\pm0.1\)    \\
DV    & \(15.1\pm1.0\)            & \(24.2\pm0.3\)            & \(57.2\pm0.3\)   \\
LG    & \(14.4\pm2.1\)            & \(8.2\pm0.5\)             & \(112.1\pm9.1\)  \\
WV    & \(12.1\pm1.1\)            & \(11.7\pm0.4\)            & \(57.2\pm0.3\)    \\
\hline                                                                          
Total & 93.2                      & 79.3                      & 321.3            \\
\hline
\end{tabular}
\label{tbl:Pascal2006}
\end{table}

\subsection{Yahoo! Front Page Click Log Dataset}

\citet{yahoo} have produced a bandit algorithm dataset. The dataset provides information about the top story presented to a user on the front page of Yahoo!. Each entry in the dataset gives information about a single article presented, the time it was presented, contextual information about the user and whether the user ``clicked-through'' to the article or not. The dataset was designed for the contextual bandit problem. Given context of a user the goal is to select an article to present to the user as to maximise the expected rate at which users click on the article to read more (click-through). The articles also change during the dataset, and so bandit algorithms designed specifically for this environment also need the ability to modify the number of arms they can select from. 
\\ 
For the purposes of our experiments we do not concern ourselves with the contextual case, nor do we try to incorporate new articles as they arrive. Instead we ignore the context, and we only pick from a set number of articles. This reduces the problem to a conventional multi-armed bandit problem. To maximise the amount of data used, for each run we randomly selected the set of articles (in our case 5 articles) from a list of \(100\) permutations of possible articles which overlapped in time the most. The click-through rates were estimated from the data by taking the mean of an articles click-through rate every \(1000\) time ticks. The simulation then proceeded as described by \citet{Li2011}, the results are presented in table \ref{tbl:yahoo}. The regret for each run was normalised by the number of arm pulls, since this was different in each run of the simulation. Parameters were set as for the PASCAL challenge dataset.
\\
\begin{table}[ht]
\centering
\scriptsize
\setlength{\tabcolsep}{2pt}
\caption{Results against Yahoo! Front Page Click Log Dataset(\(\pm\)Std. Error) }
\begin{tabular}{l r | l r }
Name & Regret & Name & Regret \\
\hline\hline
 \multicolumn{1}{|| l}{ Global-CTS    } & \(0.489\pm0.035\)  & Global-CTS2     &  \multicolumn{1}{r ||}{  \(\mathbf{0.443\pm0.031}\) }\\
 \multicolumn{1}{|| l}{ PA-CTS        } & \(0.522\pm0.028\)  & PA-CTS2         &  \multicolumn{1}{r ||}{  \(0.505\pm0.028\)          }\\
 \multicolumn{1}{|| l}{ NP Global-CTS } & \(0.490\pm0.029\)  & NP PA-CTS       &  \multicolumn{1}{r ||}{  \(0.590\pm0.018\)          }\\
 \multicolumn{1}{|| l}{ NP Global-CTS2} & \(0.530\pm0.026\)  & NP PA-CTS2      &  \multicolumn{1}{r ||}{  \(0.563\pm0.018\)          }\\
\hline
UCB            & \(0.526\pm0.040\)  & DiscountedUCB   & \(0.568\pm0.022\)          \\
Random         & \(0.800\pm0.001\)  &                 &                            \\
\end{tabular}
\label{tbl:yahoo}
\end{table}

\subsection{Foreign Exchange Rate Data}

We constructed a final test environment from Foreign Exchange Rate data\citep{dukascopy}. Ask prices for 4 currency exchange rates (GBP-USD, USD-JPY, NZD-CHF, EUR-CAD) at a resolution of 2 minutes spanning 7 years were used. This amounted to approximation \(10^6\) datapoints per exchange rate pair.  The bandit problem using this data was set up as follows. Each exchange rate was thought of as a 2-armed bandit. It was imagined that the agent could make fictitious trades, and could either decide to buy a long call option (if they believe the rate will increase) and a short call option (if they believe the rate will go down). To turn this into a Bernoulli bandit problem, we ignore the scale of the change and provide a reward of \(1\) if the bandit predicted correctly the rate going up/down and \(0\) otherwise. When the rate remains the same, the agent receives a reward of \(0\) irrespective of their decision. For the purpose of the experiment we imagine the option length is 100 time ticks, so that the agent has to decide if the exchange rate will increase or decrease in 100 time ticks.
Although this bandit scenario is not true to life, we believe that the underlying data should exhibit some of the characteristics of a switching system for which the algorithms were designed \citep{Preis2011}. We can not estimate a ``true'' average payoff at each timestep, and so can not measure the regret of these algorithms, instead we report the error. The results are shown in table \ref{tbl:Finance}.

\begin{table}[ht]
\centering
\scriptsize
\setlength{\tabcolsep}{3pt}
\caption{Results against Foreign Exchange Bandit Environment (number of mistakes \(\times 10^{-3} \pm\)Std. Error)}
\begin{tabular}{l r | l r}
Name & Error & Name & Error \\
\hline\hline
 \multicolumn{1}{|| l}{ Global-CTS    }    & \(351.9\pm14.1\)           & Global-CTS2          &  \multicolumn{1}{r ||}{  \(358.0\pm13.95\) }   \\
 \multicolumn{1}{|| l}{ PA-CTS        }    & \(370.4\pm13.7\)           & PA-CTS2              &  \multicolumn{1}{r ||}{  \(380.9\pm12.5\)  }  \\
 \multicolumn{1}{|| l}{ NP Global-CTS  }   & \(\mathbf{348.2\pm13.7}\)  & NP PA-CTS            &  \multicolumn{1}{r ||}{  \(353.5\pm13.8\)  }  \\
 \multicolumn{1}{|| l}{ NP Global-CTS2 }   & \(353.2\pm13.4\)           & NP PA-CTS2           &  \multicolumn{1}{r ||}{  \(352.0\pm13.9\)  }  \\
\hline
UCB               & \(613.9\pm17.7\)           & DiscountedUCB        & \(606.3\pm16.0\)    \\
Random            & \(623.3\pm14.1\)           &                      &                     \\
\end{tabular}
\label{tbl:Finance}
\end{table}

\section{Conclusion}
This paper has explored several algorithms using Thompson Sampling in conjunction with Change Point detection. We have shown that they perform well in the environments for which they are designed. Bandit scenarios based on real-world data such as the Yahoo! dataset and the Foreign Exchange also demonstrate their performance. 
They are shown not to perform as well as appropriately tuned competing algorithms in the PASCAL challenge. However our results suggest that a strategy that just tracks changes in the perceived best arm (Global-CTS2,PA-CTS2), similar to Adapt-EvE, works well. 
\\
Since the model is extremely modular it is hoped that further assumptions can be incorporated into the model to improve performance. It is also worth noting that non-Bernoulli payoffs can just as easily be used, e.g. Normal distributed payoffs. A Bayesian approach also avoids difficulties that arise with handling false alarms in change point detection schemes. 
The algorithms have been derived from simple models and so are theoretically motivated, however steps still need to be taken to provide any theoretic justification for their performance.

\bibliographystyle{plainnat}
\bibliography{main}

\begin{thebibliography}{14}
\providecommand{\natexlab}[1]{#1}
\providecommand{\url}[1]{\texttt{#1}}
\expandafter\ifx\csname urlstyle\endcsname\relax
  \providecommand{\doi}[1]{doi: #1}\else
  \providecommand{\doi}{doi: \begingroup \urlstyle{rm}\Url}\fi

\bibitem[Adams and MacKay(2007)]{Adams2007}
Ryan~Prescott Adams and David~J.C. MacKay.
\newblock Bayesian online changepoint detection.
\newblock Cambridge, UK, 2007.

\bibitem[Dukascopy(2012)]{dukascopy}
Dukascopy.
\newblock Dukascopy historical data feed.
\newblock \url{http://www.dukascopy.com/swiss/english/marketwatch/historical/},
  2012.
\newblock [Online; accessed 05-Oct-2012].

\bibitem[Fearnhead and Clifford(2003)]{Fearnhead2003}
Paul Fearnhead and Peter Clifford.
\newblock {On-line inference for hidden Markov models via particle filters}.
\newblock \emph{Journal of the Royal Statistical Society: Series B (Statistical
  Methodology)}, 65\penalty0 (4):\penalty0 887--899, 2003.
\newblock \doi{10.1111/1467-9868.00421}.
\newblock URL \url{http://dx.doi.org/10.1111/1467-9868.00421}.

\bibitem[Fearnhead and Liu(2007)]{Fearnhead2007}
Paul Fearnhead and Zhen Liu.
\newblock On-line inference for multiple change points problems.
\newblock \emph{Journal of the Royal Statistical Society B}, 69:\penalty0
  589--605, 2007.

\bibitem[{Garivier} and {Moulines}(2008)]{Garivier2008}
A.~{Garivier} and E.~{Moulines}.
\newblock {On Upper-Confidence Bound Policies for Non-Stationary Bandit
  Problems}.
\newblock \emph{ArXiv e-prints}, May 2008.

\bibitem[Hartland et~al.(2007)Hartland, Baskiotis, Gelly, Sebag, and
  Teytaud]{Hartland2006}
C{\'e}dric Hartland, Nicolas Baskiotis, Sylvain Gelly, Mich{\`e}le Sebag, and
  Olivier Teytaud.
\newblock {Change Point Detection and Meta-Bandits for Online Learning in
  Dynamic Environments}.
\newblock \emph{CAp}, pages 237--250, 2007.
\newblock URL \url{http://hal.inria.fr/inria-00164033}.

\bibitem[Hussain et~al.(2006)Hussain, Auer, Cesa-Bianchi, Newnham, and
  Shawe-Taylor]{PASCAL}
Z.~Hussain, P.~Auer, N.~Cesa-Bianchi, L.~Newnham, and J.~Shawe-Taylor.
\newblock Exploration vs. exploitation pascal challenge.
\newblock http://pascallin.ecs.soton.ac.uk/Challenges/EEC/, 2006.

\bibitem[Kaufmann et~al.(2012)Kaufmann, Korda, and Munos]{Kaufmann2012}
Emilie Kaufmann, Nathaniel Korda, and Rémi Munos.
\newblock Thompson sampling: An optimal finite time analysis.
\newblock \emph{CoRR}, abs/1205.4217, 2012.
\newblock URL
  \url{http://dblp.uni-trier.de/db/journals/corr/corr1205.html#abs-1205-4217}.

\bibitem[Li et~al.(2011)Li, Chu, Langford, and Wang]{Li2011}
Lihong Li, Wei Chu, John Langford, and Xuanhui Wang.
\newblock Unbiased offline evaluation of contextual-bandit-based news article
  recommendation algorithms.
\newblock In Irwin King, Wolfgang Nejdl, and Hang Li, editors, \emph{WSDM},
  pages 297--306. ACM, 2011.
\newblock ISBN 978-1-4503-0493-1.
\newblock URL
  \url{http://dblp.uni-trier.de/db/conf/wsdm/wsdm2011.html#LiCLW11}.

\bibitem[Preis et~al.(2011)Preis, Schneider, and Stanley]{Preis2011}
T.~Preis, J.~J. Schneider, and H.~E. Stanley.
\newblock {Switching processes in financial markets}.
\newblock \emph{Proceedings of the National Academy of Sciences}, 108\penalty0
  (19):\penalty0 7674--7678, April 2011.
\newblock ISSN 1091-6490.
\newblock \doi{10.1073/pnas.1019484108}.
\newblock URL \url{http://dx.doi.org/10.1073/pnas.1019484108}.

\bibitem[Sorensen(2007)]{Sorensen2007}
Morten Sorensen.
\newblock Learning by investing: Evidence from venture capital.
\newblock SIFR Research Report Series~53, Institute for Financial Research, May
  2007.
\newblock URL \url{http://ideas.repec.org/p/hhs/sifrwp/0053.html}.

\bibitem[Turner et~al.(2009)Turner, Saatci, and Rasmussen]{Turner2009}
Ryan Turner, Yunus Saatci, and Carl~Edward Rasmussen.
\newblock Adaptive sequential {B}ayesian change point detection.
\newblock In \emph{Advances in Neural Information Processing Systems (NIPS):
  Temporal Segmentation Workshop}, 2009.

\bibitem[Wilson et~al.(2010)Wilson, Nassar, and Gold]{Wilson2010}
Robert~C. Wilson, Matthew~R. Nassar, and Joshua~I. Gold.
\newblock Bayesian online learning of the hazard rate in change-point problems.
\newblock \emph{Neural Comput.}, 22\penalty0 (9):\penalty0 2452--2476, 2010.
\newblock ISSN 0899-7667.
\newblock \doi{10.1162/NECO_a_00007}.
\newblock URL \url{http://dx.doi.org/10.1162/NECO_a_00007}.

\bibitem[Yahoo!(2011)]{yahoo}
Yahoo!
\newblock Yahoo! webscope dataset ydata-frontpage-todaymodule-clicks-v1\_0.
\newblock \url{http://labs.yahoo.com/Academic_Relations}, 2011.
\newblock [Online; accessed 05-Oct-2012].

\end{thebibliography}

\end{document}